\documentclass[letterpaper, 10 pt, conference]{ieeeconf}  

\IEEEoverridecommandlockouts                              

\usepackage{graphicx}
\usepackage{amsmath}
\usepackage{booktabs}
\usepackage{multirow}
\usepackage{amsmath} 
\usepackage{subcaption}  
\usepackage{float}       
\usepackage{nicefrac}       
\usepackage{xcolor}

\definecolor{tuebingenred}{RGB}{165, 30, 55}

\title{\LARGE \bf
Lightweight Multi-Frame Integration for 

Robust YOLO Object Detection in Videos
}

\author{Yitong Quan$^{1}$, Benjamin Kiefer$^{1}$, Martin Messmer$^{1}$ and Andreas Zell$^{1}$
\thanks{$^{1}$University of Tuebingen 
        {\tt\small first.last@uni-tuebingen.de}}%
}

\begin{document}

\maketitle
\thispagestyle{empty}
\pagestyle{empty}

\begin{abstract}

Modern image-based object detection models, such as YOLOv7, primarily process individual frames independently, thus ignoring valuable temporal context naturally present in videos. Meanwhile, existing video-based detection methods often introduce complex temporal modules, significantly increasing model size and computational complexity. In practical applications such as surveillance and autonomous driving, transient challenges including motion blur, occlusions, and abrupt appearance changes can severely degrade single-frame detection performance. To address these issues, we propose a straightforward yet highly effective strategy: stacking multiple consecutive frames as input to a YOLO-based detector while supervising only the output corresponding to a single target frame. This approach leverages temporal information with minimal modifications to existing architectures, preserving simplicity, computational efficiency, and real-time inference capability. Extensive experiments on the challenging MOT20Det and our BOAT360 datasets demonstrate that our method  improves detection robustness, especially for lightweight models, effectively narrowing the gap between compact and heavy detection networks. Additionally, we contribute the BOAT360 benchmark dataset, comprising annotated fisheye video sequences captured from a boat, to support future research in multi-frame video object detection in challenging real-world scenarios.

\end{abstract}

\section{INTRODUCTION}
Modern object detection methods, notably YOLO-based architectures such as YOLOv7~\cite{wang2023yolov7}, have achieved remarkable performance on single-frame image benchmarks. However, when deployed in video-centric applications such as robotics, surveillance systems, and autonomous driving, these methods often fall short due to their limited ability to utilize temporal information. Single-frame detectors typically treat each video frame independently, failing to exploit valuable cues from preceding frames, thereby reducing robustness to real-world challenges including occlusions, motion blur, and rapid changes in appearance.

\begin{figure}[h]
    \centering
    \begin{subfigure}[b]{0.48\textwidth}
        \includegraphics[width=\textwidth]{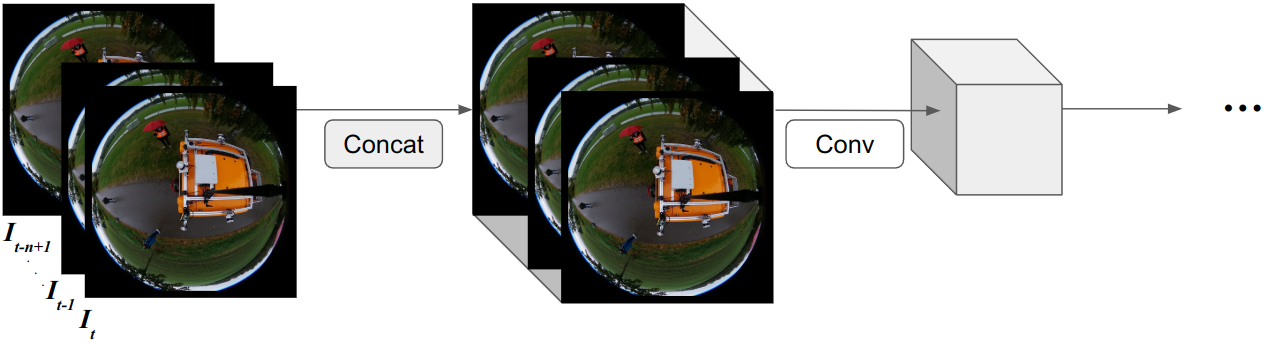}
        \caption{Early Fusion (Standard Convolution)}
    \end{subfigure}
    \hfill
    \begin{subfigure}[b]{0.48\textwidth}
        \includegraphics[width=\textwidth]{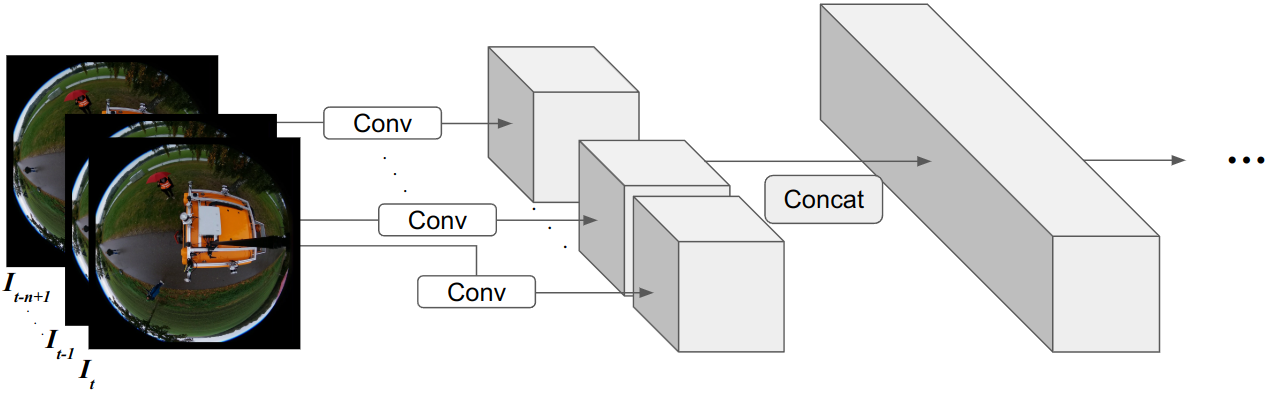}
        \caption{ Late Fusion (Grouped Convolution)}
    \end{subfigure}
    \caption{Network adaptation strategies for multi-frame object detection. We investigate two methods to extend single-frame detectors to support multiple video frames as input. (Top) Early Fusion (Standard Convolution): Input frames are stacked along the channel dimension and passed through shared convolutions from the first layer. This allows the network to jointly learn spatiotemporal features directly from raw pixels. (Bottom) Late Fusion (Grouped Convolution): Each frame is independently processed by separate convolutional kernels in a grouped convolution setup. Early layers extract low-level features per frame before merging via concatenation. This defers temporal fusion until later in the network. Both methods require modifying only the first few layers, making them lightweight and compatible with existing pretrained models.}
    \label{fig:Network_adaptation}
\end{figure}

To address this limitation, we propose a simple yet effective strategy: feeding a stack of multiple consecutive frames into the detection network and supervising the detection output for only the most recent frame. This approach implicitly enables the network to learn temporal and motion cues directly from raw pixel data at the earliest stage of processing, potentially enhancing the robustness of object detection without significantly increasing the model's complexity or the training overhead.

Unlike previous approaches that explicitly model temporal dynamics using additional computationally intensive modules such as recurrent networks~\cite{ahmad2020recurrent}, optical flow estimation~\cite{zhu2017flow}, or temporal attention mechanisms~\cite{wu2019sequence,feichtenhofer2019slowfast}, our proposed method integrates seamlessly into existing single-frame architectures with minimal adaptation. Importantly, it does not need dense labels for each frame as opposed to classical video object detection methods. It merely takes in preceding video frames without labels as additional context, effectively categorizing our approach into a sparsely or \emph{weakly supervised} learning method \cite{zhou2018brief}.

Specifically, we extend YOLOv7 and YOLOv7-tiny by adapting the first convolutional layer to accommodate multiple input frames. We explore two convolutional adaptation strategies: direct early fusion by stacking frames at the pixel level, and grouped convolutions where each frame is processed independently before concatenation at the feature level.

We  evaluate our method on the MOT20Det dataset \cite{dendorfer2020mot20}, a challenging pedestrian detection benchmark known for its crowded environments and frequent occlusions. Our results  demonstrate that integrating temporal information  boosts the detection performance of lightweight models such as YOLOv7-tiny. Furthermore, we validate the generalizability of our method on our BOAT360 dataset, which involves dynamic scenarios captured by a moving fisheye camera, showcasing our method's robustness and wide applicability.


In summary, our key contributions are as follows: \begin{itemize} \item \textbf{A lightweight yet highly effective temporal integration strategy for YOLO-based object detectors}, which improves detection robustness while maintaining computational efficiency and minimal architectural modifications. \item \textbf{Extensive empirical validation}, demonstrating performance gains over single-frame YOLO baselines across challenging datasets, including MOT20Det and our own BOAT360 dataset, under conditions of occlusion, motion blur, and rapid appearance changes. \item \textbf{Benchmark dataset contribution}, providing the BOAT360 dataset featuring annotated fisheye video sequences captured from a dynamic, waterborne platform, to support further research in multi-frame video object detection dynamic real-world environments upon paper acceptance. \end{itemize}

\section{RELATED WORK}

\subsection{Single-Frame Object Detection}

Object detection has witnessed tremendous progress with deep convolutional networks. Architectures like Faster R-CNN \cite{ren2015fasterrcnn}, YOLO \cite{redmon2016you,wang2023yolov7}, and RetinaNet \cite{lin2017focal} have demonstrated strong performance on still images. Recent models such as  YOLOv7 \cite{wang2023yolov7} further optimize the trade-off between accuracy and speed, enabling real-time performance on edge devices.

However, these models primarily process each frame independently and do not exploit the temporal consistency naturally available in videos. In contrast, our method integrates multiple consecutive frames as input, providing temporal context while maintaining the architectural simplicity and speed advantages of YOLO-based detectors.

\subsection{Video Object Detection}

Video object detection aims to improve detection robustness by utilizing temporal information across frames. Early works proposed simple techniques such as feature aggregation across adjacent frames \cite{zhu2017flow} or tracking-by-detection pipelines \cite{stadler2023detailed}. More recent approaches introduced flow-guided feature warping \cite{zhu2017flow}, temporal attention \cite{wu2020video}, or recurrent memory networks \cite{pu2020learning}.

While these methods achieve notable gains, they often require complex architectures or multi-stage training procedures, making them difficult to integrate into real-time systems.
Our approach offers simpler alternatives without adding large temporal modeling modules, making it more suitable for lightweight and deployable applications.

\subsection{Multi-Frame Input with Sparse Supervision}

Several studies have explored feeding multiple frames into the detector directly. Simple early fusion by stacking frames was explored in \cite{qu2022video}, but often limited to two frames or relied on heavy video models such as 3D CNNs \cite{hou2019efficient}. 

Moreover, while most methods supervise every (or almost every) frame in a video sequence \cite{jiao2021new,wu2023label,zhu2020review,shi2023yolov}, we propose to supervise only a single frame while still benefiting from multiple inputs. This reduces labeling dependency and fits well into settings where annotation density is limited. In fact, many proposed object detection datasets sprung from videos where most frames were discarded in favor of diverse scenes \cite{varga2022seadronessee,zhu2020review,kiefer20232nd}. Our approach allows to leverage these datasets to better performance almost \emph{for free}.

\subsection{Model Interpretability in Object Detection}

Recent interpretability tools such as Grad-CAM~\cite{selvaraju2017grad}, Grad-CAM++~\cite{chattopadhay2018grad} and Eigen-CAM~\cite{muhammad2020eigen} help visualize which regions influence predictions. In our study, we apply Grad-CAM++ to qualitatively analyze where the YOLO detector focuses when using single vs. multi-frame inputs.

\section{METHODOLOGY}

\subsection{Problem Formulation}

Given a video sequence, our goal is to improve object detection by leveraging information from multiple consecutive frames, while still supervising the model only on a single frame per input stack. This setting reduces annotation costs and introduces temporal context that aids detection under occlusion, motion blur, and appearance changes.

Formally, let $\{I_t\}_{t=1}^T$ be a sequence of RGB frames. Instead of detecting objects based only on frame $I_t$, we provide the model with a stack of $n$ frames $\{I_{t-n+1}, \dots, I_{t-1}, I_t\}$ and predict object detections for frame $I_t$ only.

\subsection{Multi-Frame Input Representation}

Instead of introducing new temporal modules or recurrent structures, we simply concatenate $n$ consecutive RGB frames along the channel dimension.  
For instance, if $n=5$, the input tensor has $3 \times n = 15$ channels, shaped as $(15, H, W)$.

The frames are stacked either as:
\begin{itemize}
    \item \textbf{Adjacent frames:} consecutive frames at native frame rate (e.g., 25fps in MOT20Det, 20fps in BOAT360).
    \item \textbf{Stepped frames:} frames sampled with a fixed gap to span a longer temporal window.
\end{itemize}

This simple early fusion exposes the model to temporal variations at the pixel level from the first convolutional layer onward.

\subsection{Network Architecture Adaptation}

We adopt YOLOv7-tiny and YOLOv7 \cite{wang2023yolov7} as our base architectures. To accommodate the increased number of input channels introduced by stacking multiple frames, we modify the first convolutional layer and explore two strategies:

\begin{itemize}
    \item \textbf{Early Fusion (Standard Convolution):} Frames are stacked along the channel dimension, and a standard convolution is applied jointly across all channels. This enables the model to integrate spatial and temporal information from the very first layer. 
    
    To leverage pretraining, the weights of the first convolutional layer are initialized by repeating the pretrained single-frame weights $n$ times along the input channel dimension and scaling them by a factor of $1/n$. This initialization ensures that, at the start of training, the response of the first convolution remains consistent with the original single-frame model, facilitating stable optimization and faster convergence.

    \item \textbf{Grouped Convolution (Feature-level Fusion):} The first convolutional layer is restructured as a grouped convolution \cite{krizhevsky2017imagenet} with $n$ groups, where each group independently processes the channels of one input frame. After this initial independent feature extraction, the feature maps from all frames are concatenated along the channel dimension. 
    
    To transfer knowledge from the pretrained single-frame model, each group is initialized with a copy of the original first-layer weights. Subsequently, the second convolutional layer is adjusted in the same way as in early fusion, ensuring compatibility with the expanded number of feature channels.
\end{itemize}

To provide a concise comparison of these strategies, Table~\ref{tab:arch_1st_layer} illustrates an example modification made to the first convolutional layer when adapting the network for multi-frame input.

\begin{table}[ht]
\centering
\caption{Adaptation of the first convolutional layer for multi-frame input. The table compares the original single-frame design with two multi-frame approaches: EF Multi (Early Fusion) and GC Multi (Grouped Convolution). We assume $n$ RGB frames of size $640 \times 640$ are input. N=32 kernels and stride of 2 are applied.}
\label{tab:arch_1st_layer}
\begin{tabular}{lccc}
\toprule
\textbf{Model} & \textbf{Kernel Size}& \textbf{Output Size} & \textbf{Weight Shape} \\
 & H, W, C& H, W, C&N, H, W, C\\
\midrule
Single frame       & $3, 3,3 $& $320,320,32$& $32,3,3,3$\\
EF Multi& $3, 3, 3n$& $320,320,32$& $32,3,3,3n$\\
GC Multi&  $3, 3,3 $ ($n$ groups) & $320,320,32n$& $32,3,3,3n$\\
\bottomrule
\end{tabular}
\end{table}





\subsection{Training Strategy}

\begin{figure}[t]
    \centering
    \includegraphics[width=0.9\linewidth]{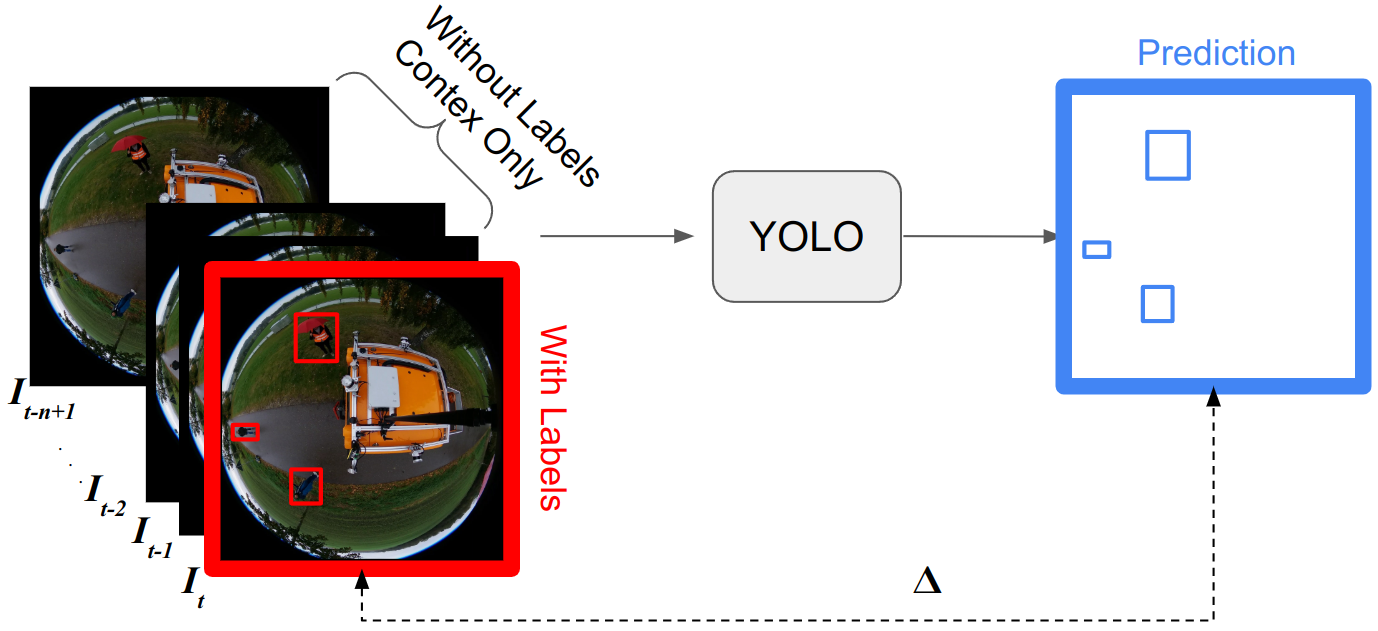}
    \caption{
    Overview of the weakly supervised training setup. Multiple frames are stacked and passed into the YOLO model, but only the latest frame \( I_t \) (marked as red) carries supervision. The earlier frames serve as unlabeled context, helping the model extract temporal cues without requiring additional labels. The sparsity of labeled frames can be adapted to the application.
    }
    \label{fig:weakly_supervised}
\end{figure}

A key advantage of our framework is its weak supervision setup: although each input consists of a stack of multiple frames, only the most recent frame \( I_t \) in the sequence is associated with ground-truth annotations. As illustrated in Figure~\ref{fig:weakly_supervised}, the preceding unlabeled frames are treated as temporal context, allowing the model to extract motion patterns and temporal dependencies across time. This design significantly reduces labeling requirements, as it eliminates the need to annotate every frame in a sequence. Our approach is well-suited for scenarios where dense frame-wise annotation is impractical or costly, such as long-duration surveillance footage or onboard recordings from mobile platforms.

During training, the input is a stack of $n$ frames. 
Augmentations such as scaling, translation, rotation, and cropping, which alter the spatial characteristics of the image, are applied consistently across all frames within each stack to maintain spatial alignment. 



Standard YOLO loss functions for objectness, classification, and bounding box regression are applied. All configurations were trained for  500 epochs, with the best checkpoint (all reached before the final epoch) selected for testing.

\subsection{Advantages of the Proposed Approach}

Our method offers:
\begin{itemize}
    \item \textbf{Simplicity:} Only minor changes to existing models.
    \item \textbf{Efficiency:} Minimal overhead in parameters and computation.
    \item \textbf{Sparse supervision:} Only one frame per stack needs to be labeled.
\end{itemize}

We demonstrate in experiments that this approach improves detection robustness, particularly in challenging scenes with occlusions and motion artifacts.

\begin{table*}[ht]
\centering
\caption{Dataset Characteristics}
\label{tab:dataset_comparison}
\begin{tabular*}{\textwidth}{@{\extracolsep{\fill}} ccccc c c }
\toprule
Dataset & Platform & Camera Mounting & Camera Type & Environment & Image Resolution &  \# Box\\ 
\midrule
MOT20Det & Static & Fixed height  & Perspective & Indoor \& Outdoor & downscaled to 640 × 640 & 2M \\
BOAT360 & Mobile (boat) & Mounted on a mast & Fisheye & Open water & downscaled to 1452 × 1452 & 509\\
\bottomrule
\end{tabular*}
\end{table*}

\section{EXPERIMENTS}

\subsection{Datasets}
We evaluate our proposed multi-frame object detection approach using two distinct datasets. These datasets differ in terms of camera setup, environmental conditions, and the complexity of the scenes, providing a robust basis for testing the generalization and performance of our method. The datasets are summarized in Table \ref{tab:dataset_comparison}.

\begin{itemize}
    \item \textbf{MOT20Det (public dataset)}: A large-scale pedestrian detection dataset captured from multiple static surveillance cameras under crowded conditions. Each annotated object includes tight bounding boxes around visible pedestrians, providing a challenging benchmark for object detection, especially under heavy occlusions and dense environments \cite{dendorfer2020mot20}.
    

    \item \textbf{BOAT360 (our dataset)}: Captured from a moving boat with a mast-mounted fisheye camera. Annotated objects are floating buoys on water from varying distances. Due to the constant motion of the boat and the unstable environment, BOAT360 offers a unique perspective for evaluating detection models for small objects under dynamic  and unstructured  background and camera movement scenarios. 
\end{itemize}

Each dataset features varying camera motion characteristics, object densities, and environmental conditions, making them ideal for evaluating the robustness and generalization capabilities of our proposed multi-frame detection method. 

As the MOT20Det dataset only provides designated training and testing sets, we further split the official training sequences into 4/5 for training and 1/5 for validation. The official testing sequences are reserved exclusively for final evaluation. 
For the BOAT360 dataset, we divide the collected video sequences into 3/5 for training, 1/5 for validation, and 1/5 for testing.






\subsection{Evaluation Metrics}
We evaluate detection performance using standard COCO-style metrics \cite{lin2014microsoft}, including Precision (P), Recall (R), mean Average Precision at an IoU threshold of 0.5 (mAP@0.5), and mean Average Precision averaged over multiple IoU thresholds from 0.5 to 0.95 in steps of 0.05 (mAP@0.5:0.95).

We also report  model parameters, floating point operations (FLOPS) on NVIDIA RTX3090 and inference speed on Orin AGX embedded GPU.


\subsection{Results}

We evaluate our multi-frame detection framework on the MOT20Det dataset and validate its generalization on the BOAT360 dataset. 

\subsubsection{Effect of Temporal Context}
As shown in Table~\ref{tab:adjacent_frames}, 
introducing multiple frames substantially improves detection performance. Stacking three adjacent frames increases mAP@0.5 by 4.4 points over the single-frame baseline. Performance further improves with 5 and 7 frames, peaking at 85.5\% mAP@0.5 and 47.8\% mAP@0.5:0.95. 

However, using 9 adjacent frames degrades precision and overall mAP, suggesting that excessively large temporal windows introduce noise rather than useful motion cues. This points to an optimal range of temporal context for balancing new information and stability.

\begin{table}[ht]
\centering
\caption{Impact of number of adjacent frames on detection performance on MOT20Det test set.}
\label{tab:adjacent_frames}
\begin{tabular}{lcccc}
\toprule
\textbf{YOLOv7-tiny} & \textbf{P} & \textbf{R} & \textbf{mAP@0.5} & \textbf{mAP@.5:.95} \\
\midrule
\textit{1 frame (baseline)} & 0.877 & 0.720 & 0.797 & 0.413 \\
3 frames adjacent & 0.943 & 0.731 & 0.841 & 0.462 \\
5 frames adjacent & 0.920 & 0.725 & 0.819 & 0.450 \\
7 frames adjacent & \textbf{0.946} & \textbf{0.755} & \textbf{0.855} & \textbf{0.478} \\
9 frames adjacent & 0.906 & 0.743 & 0.829 & 0.447 \\
\bottomrule
\end{tabular}
\end{table}

\subsubsection{Sparse Temporal Sampling}

Since 7 adjacent frames yielded the best accuracy in dense settings (Table~\ref{tab:adjacent_frames}), we explore whether similar temporal coverage with fewer, more widely spaced frames could achieve comparable results. As shown in Table~\ref{tab:temporal_steps}, 3 frames with a step of 3 (spanning 6 frames) and 5 frames with a step of 2 (spanning 8-frame duration) both match or outperform the 7-frame adjacent baseline in mAP and recall. 
Additionally, a dynamic sampling configuration  that emphasizes recent frames while retaining long context performs competitively. Overall, sparse temporal sampling offers an effective and efficient alternative to dense stacking, balancing temporal coverage and input efficiency.



\begin{table}[ht]
\centering
\caption{Impact of temporal step sampling on detection performance on MOT20Det.}
\label{tab:temporal_steps}
\begin{tabular}{lcccc}
\toprule
\textbf{YOLOv7-tiny} & \textbf{P} & \textbf{R} & \textbf{mAP@0.5} & \textbf{mAP@.5:.95} \\
\midrule
7 frames adjacent  & \textbf{0.946} & 0.755 & 0.855 & 0.478 \\
3 frames, step 3 & 0.926 & \textbf{0.802} & \textbf{0.873} & \textbf{0.490} \\
4 frames, step 2 & 0.948 & 0.736 & 0.844 & 0.466 \\
5 frames, step 2 & 0.939 & 0.782 & 0.870 & \textbf{0.490} \\
4 f., index [-6,-3,-1,0] & 0.947 & 0.753 & 0.852 & 0.470 \\
\bottomrule
\end{tabular}
\end{table}

\begin{figure*}[t]
    \centering
    \setlength{\tabcolsep}{1pt}
    \renewcommand{\arraystretch}{0.5}
    \begin{tabular}{cccc}
        \includegraphics[width=0.24\textwidth]{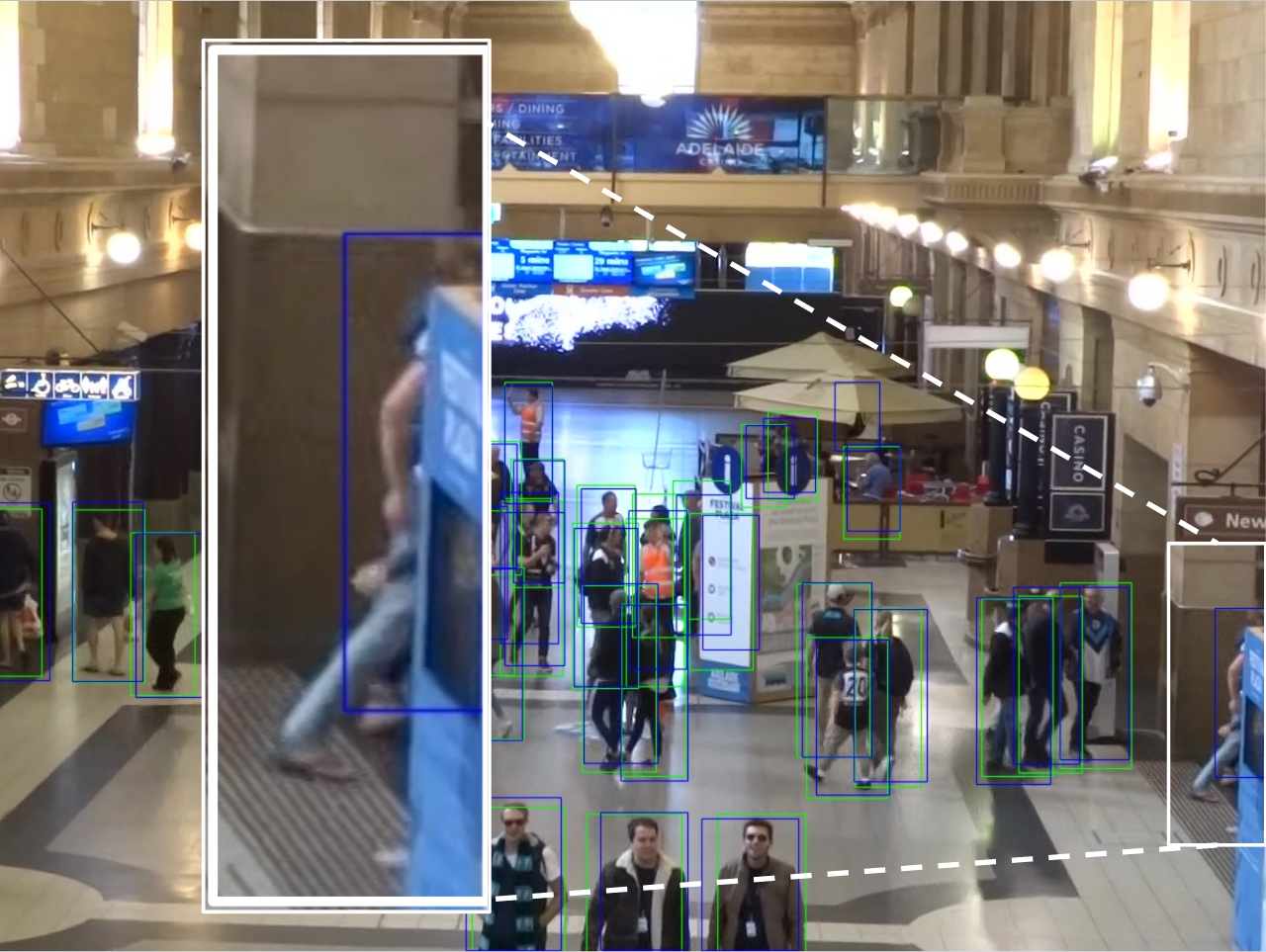} &
        \includegraphics[width=0.24\textwidth]{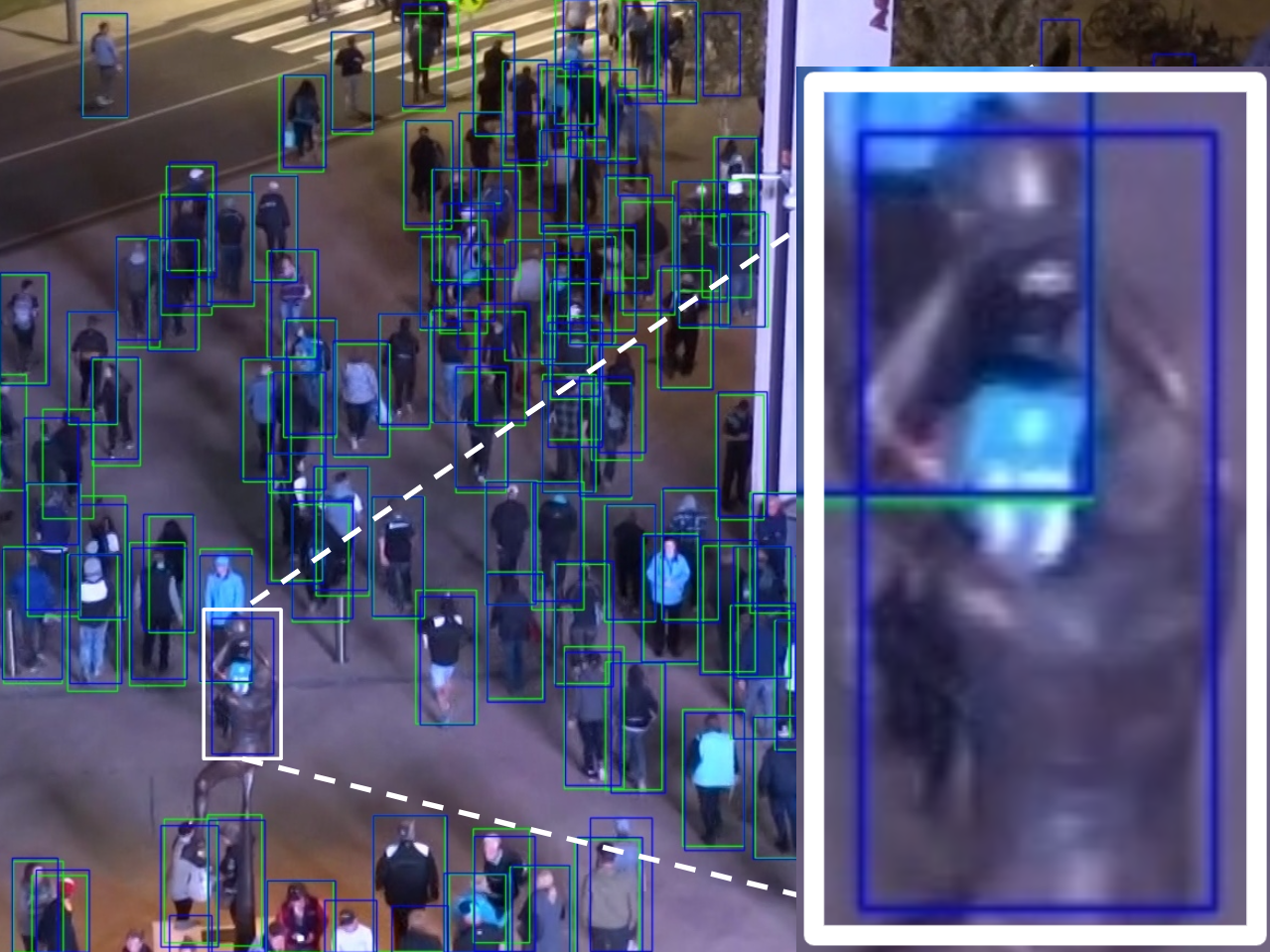} &
        \includegraphics[width=0.24\textwidth]{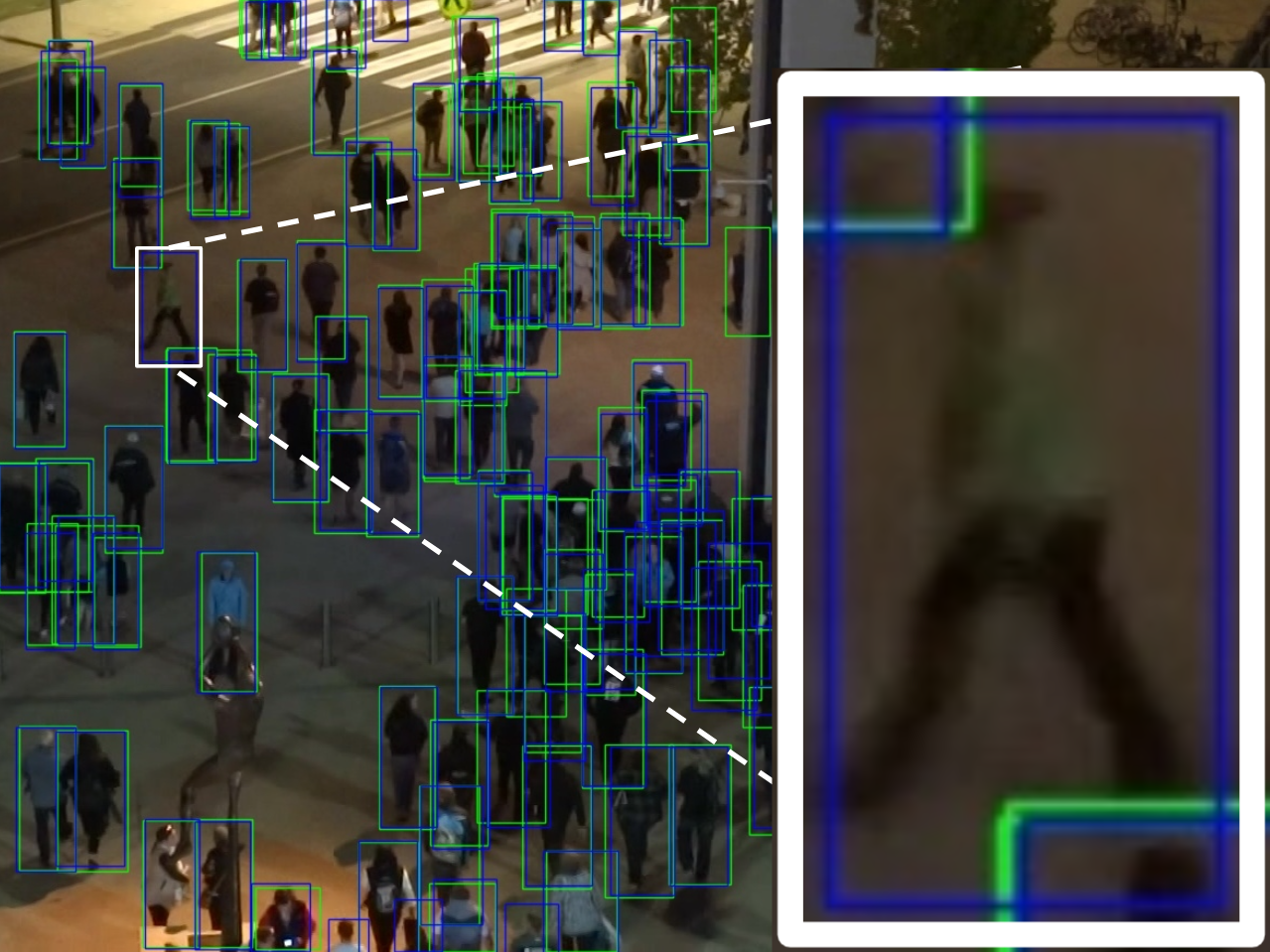} &
        \includegraphics[width=0.24\textwidth]{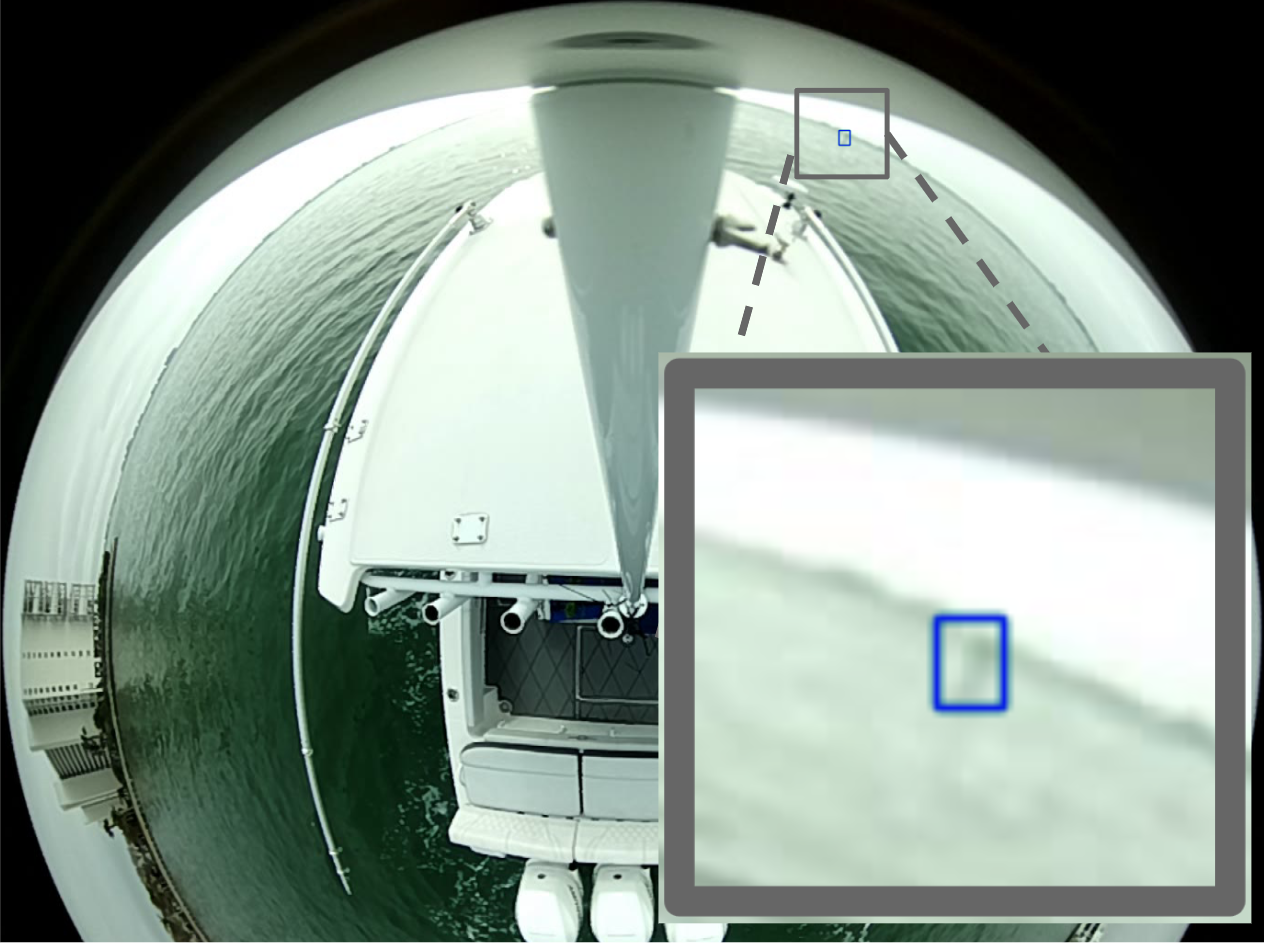} \\

        \includegraphics[width=0.24\textwidth, trim=16.95cm 0 0 0, clip]{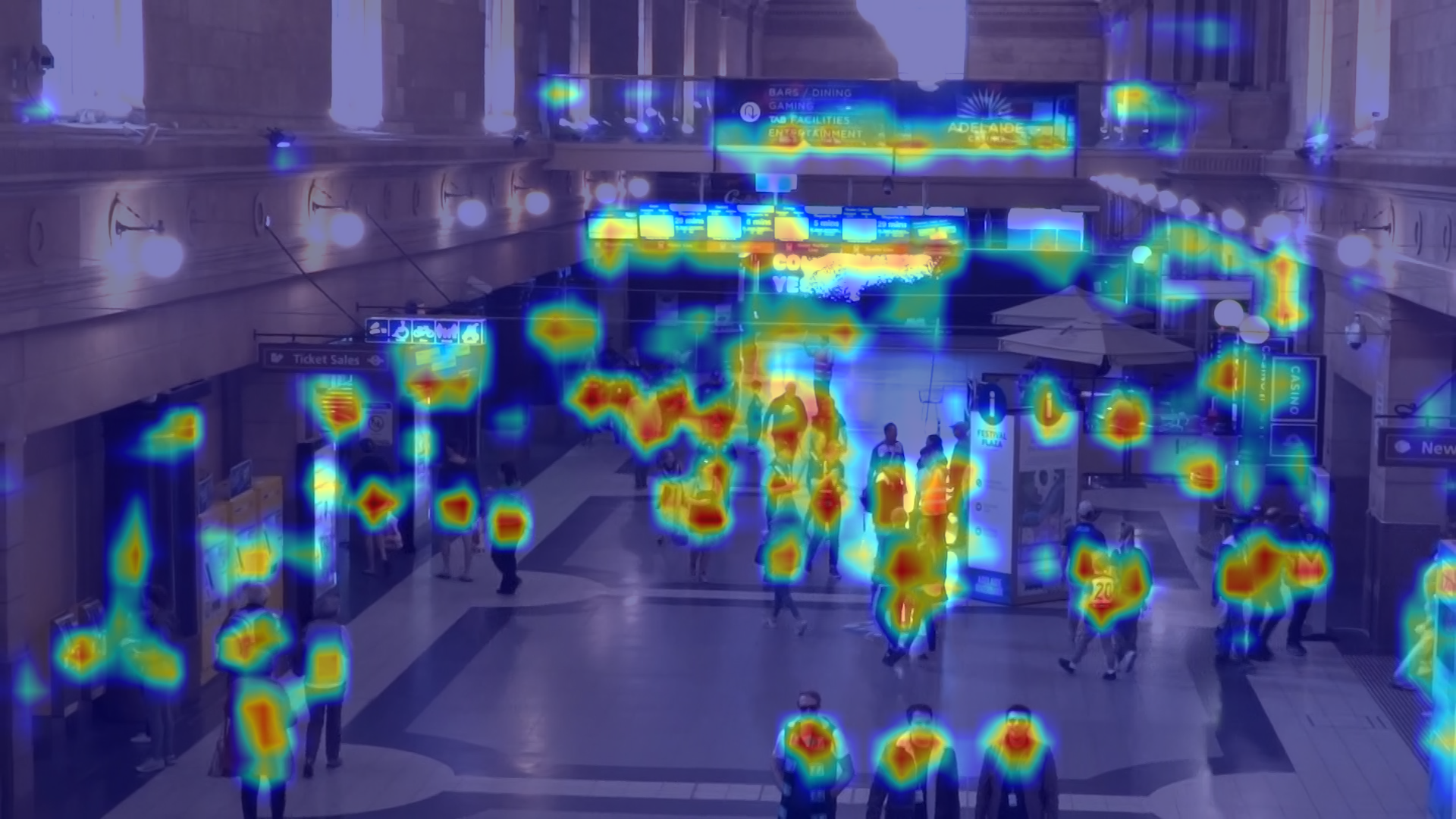} &
        \includegraphics[width=0.24\textwidth]{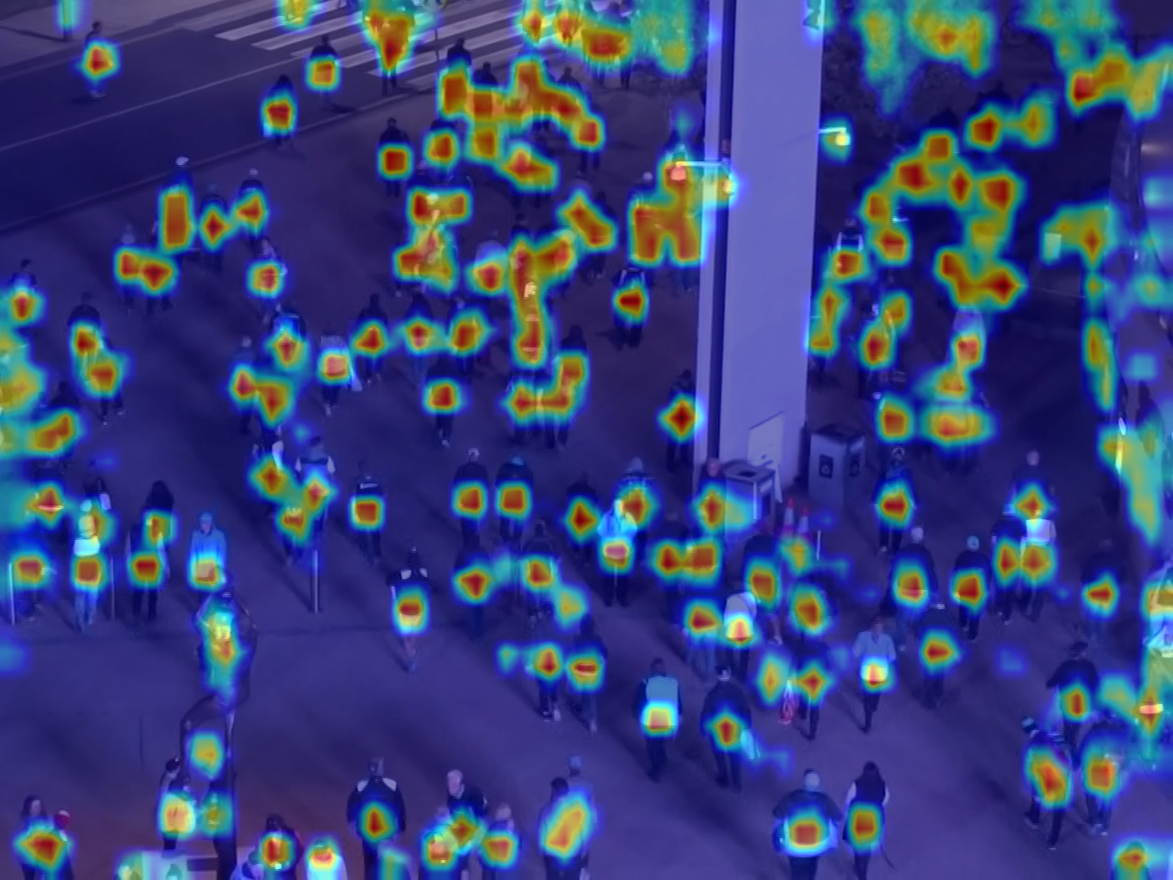} &
        \includegraphics[width=0.24\textwidth]{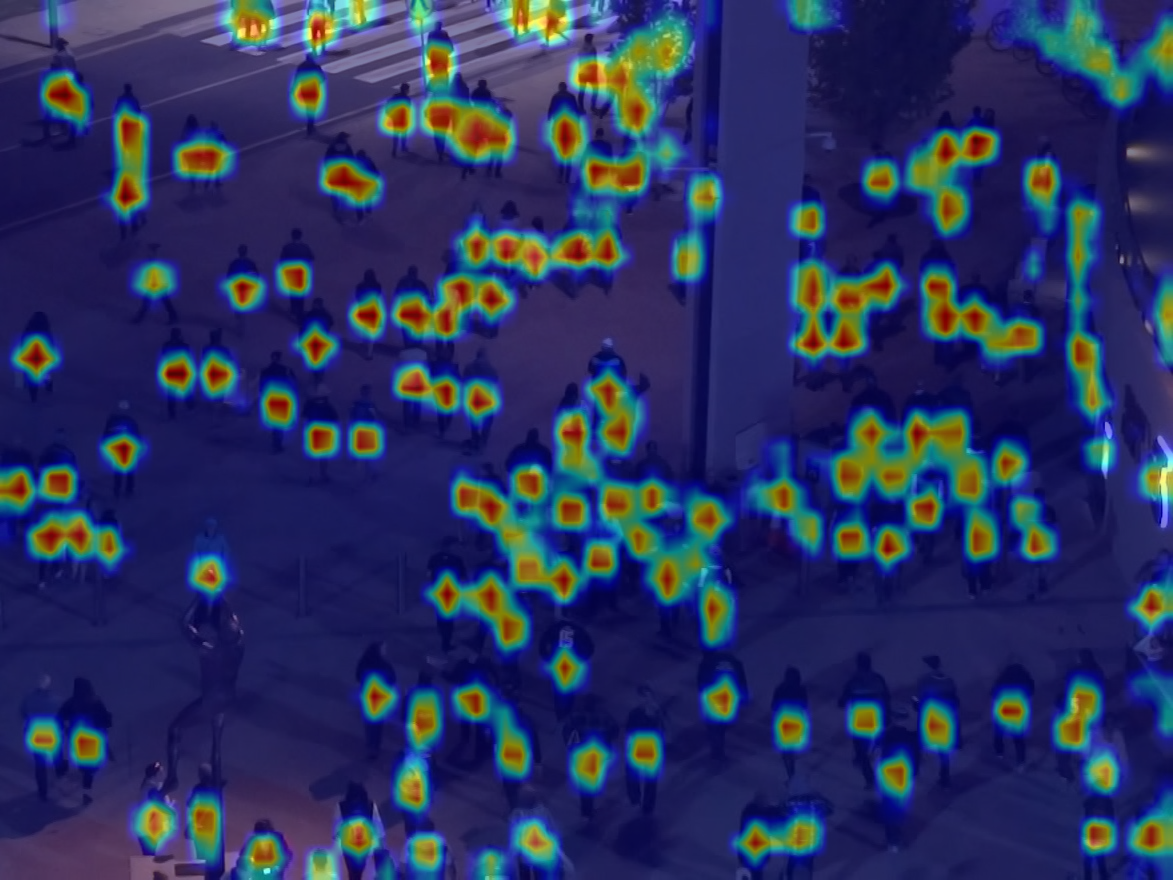} &
        \includegraphics[width=0.24\textwidth, trim=0 12.05cm 0 0, clip]{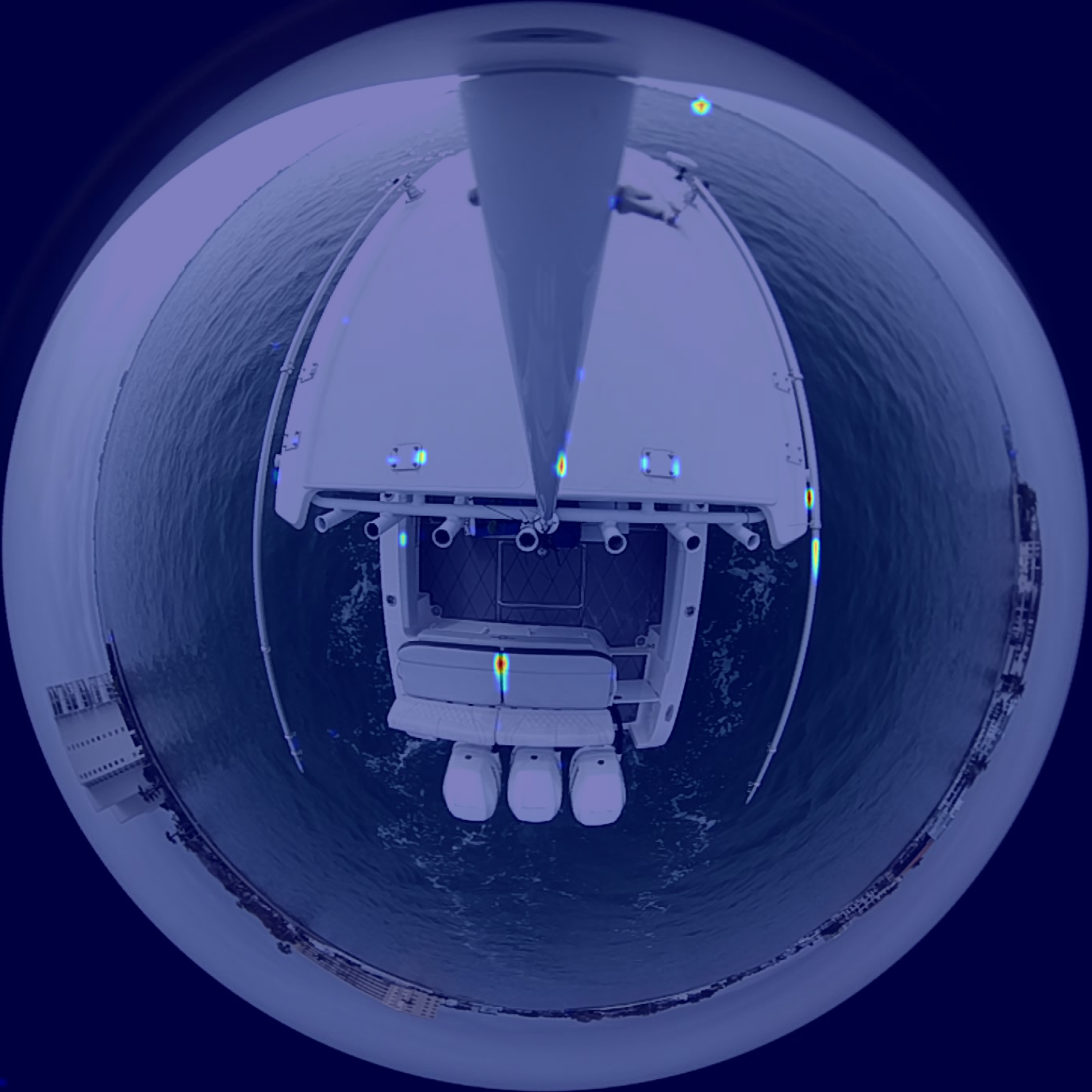} \\

        \includegraphics[width=0.24\textwidth, trim=16.95cm 0 0 0, clip]{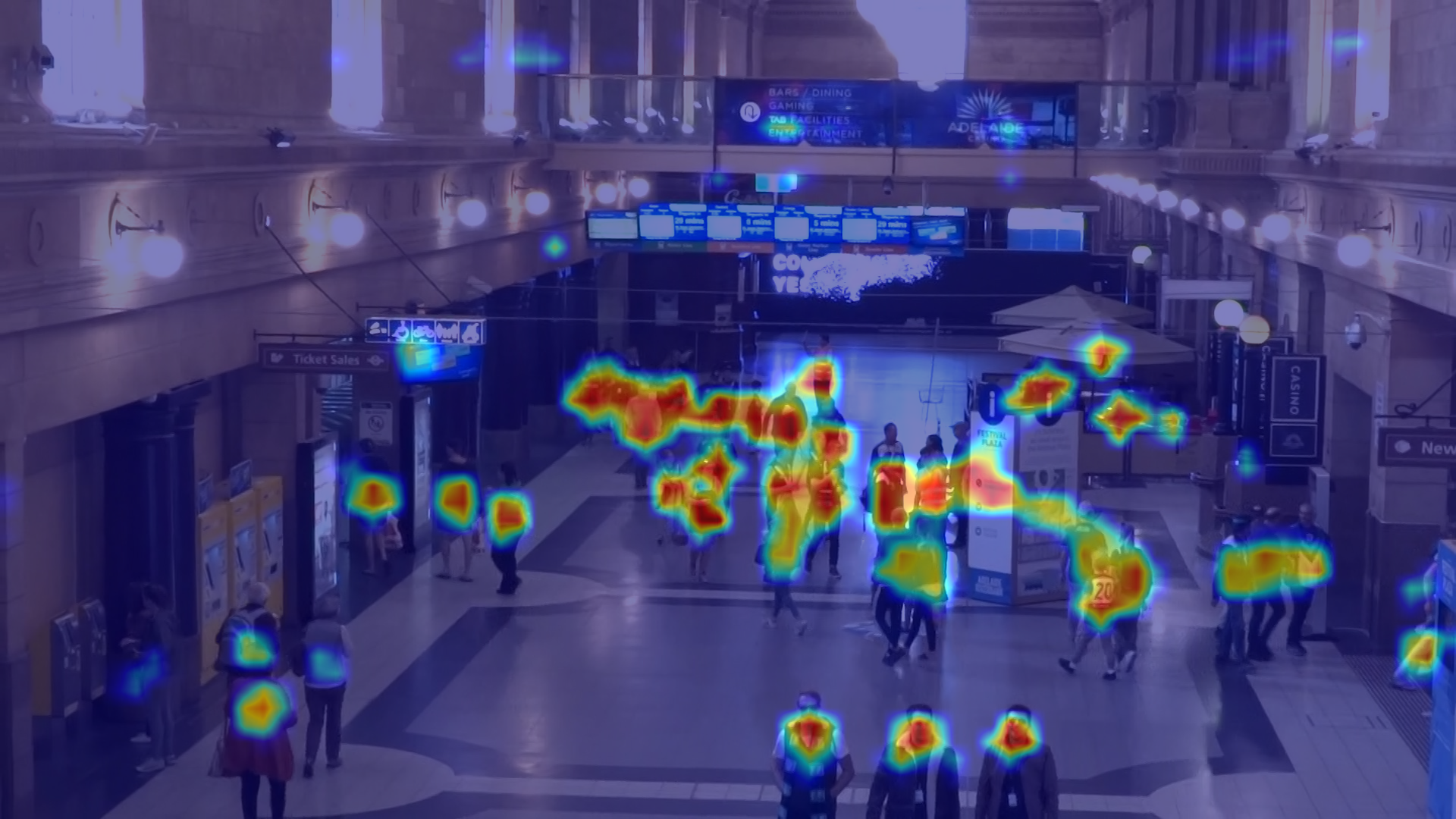} &
        \includegraphics[width=0.24\textwidth]{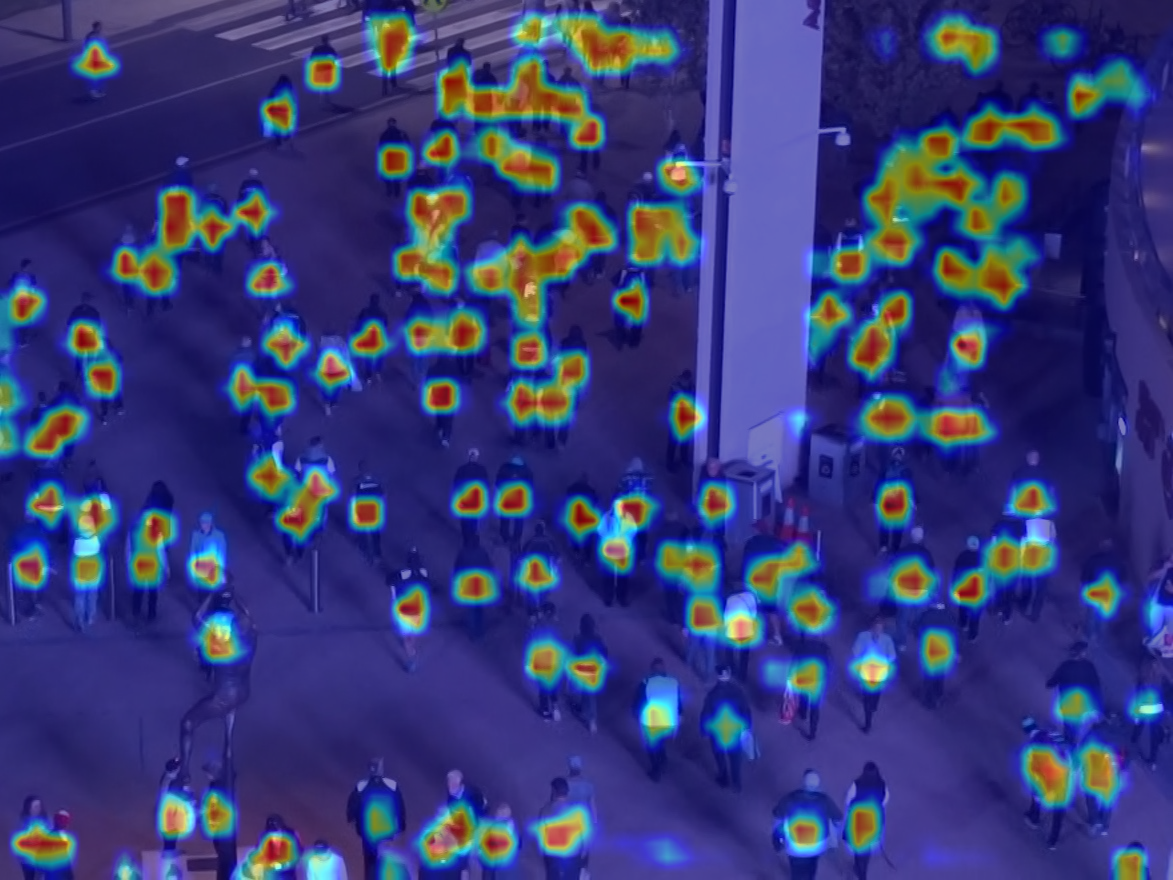} &
        \includegraphics[width=0.24\textwidth]{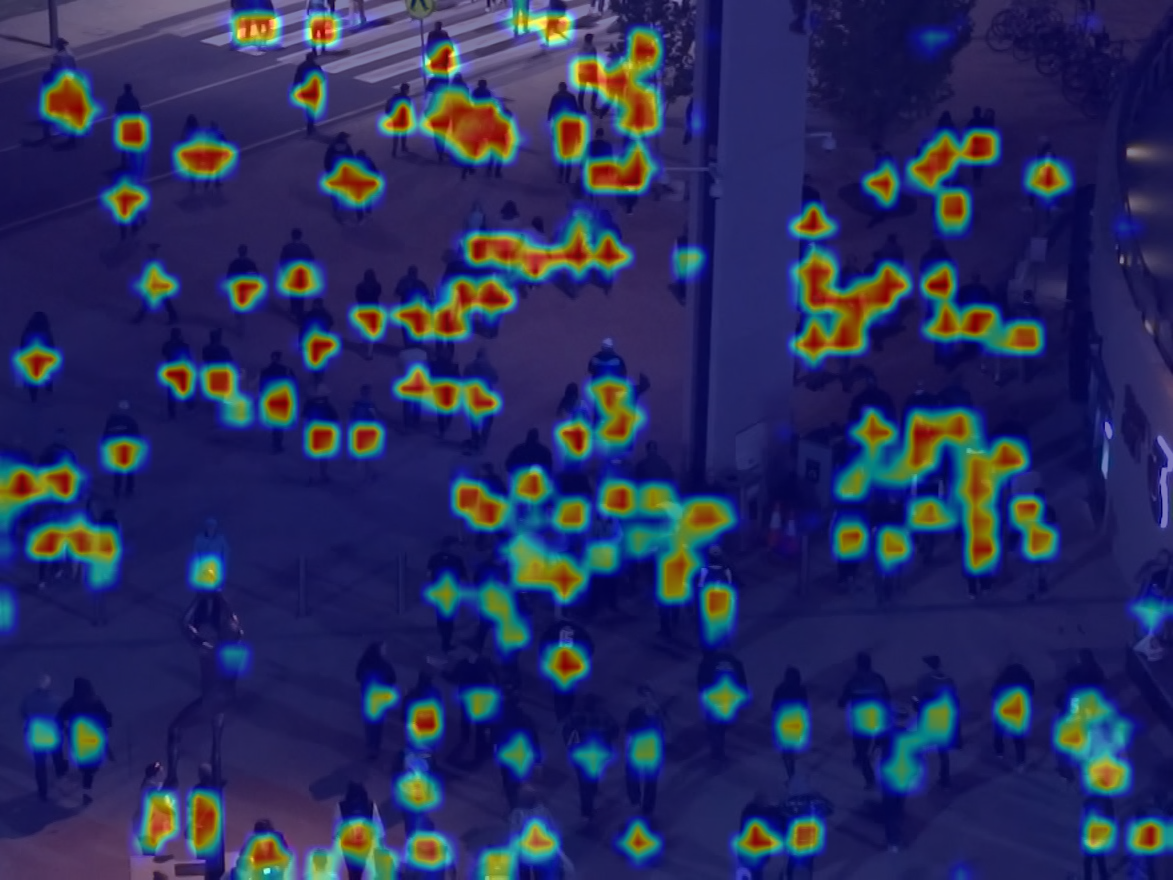} &
        \includegraphics[width=0.24\textwidth, trim=0 12.05cm 0 0, clip]{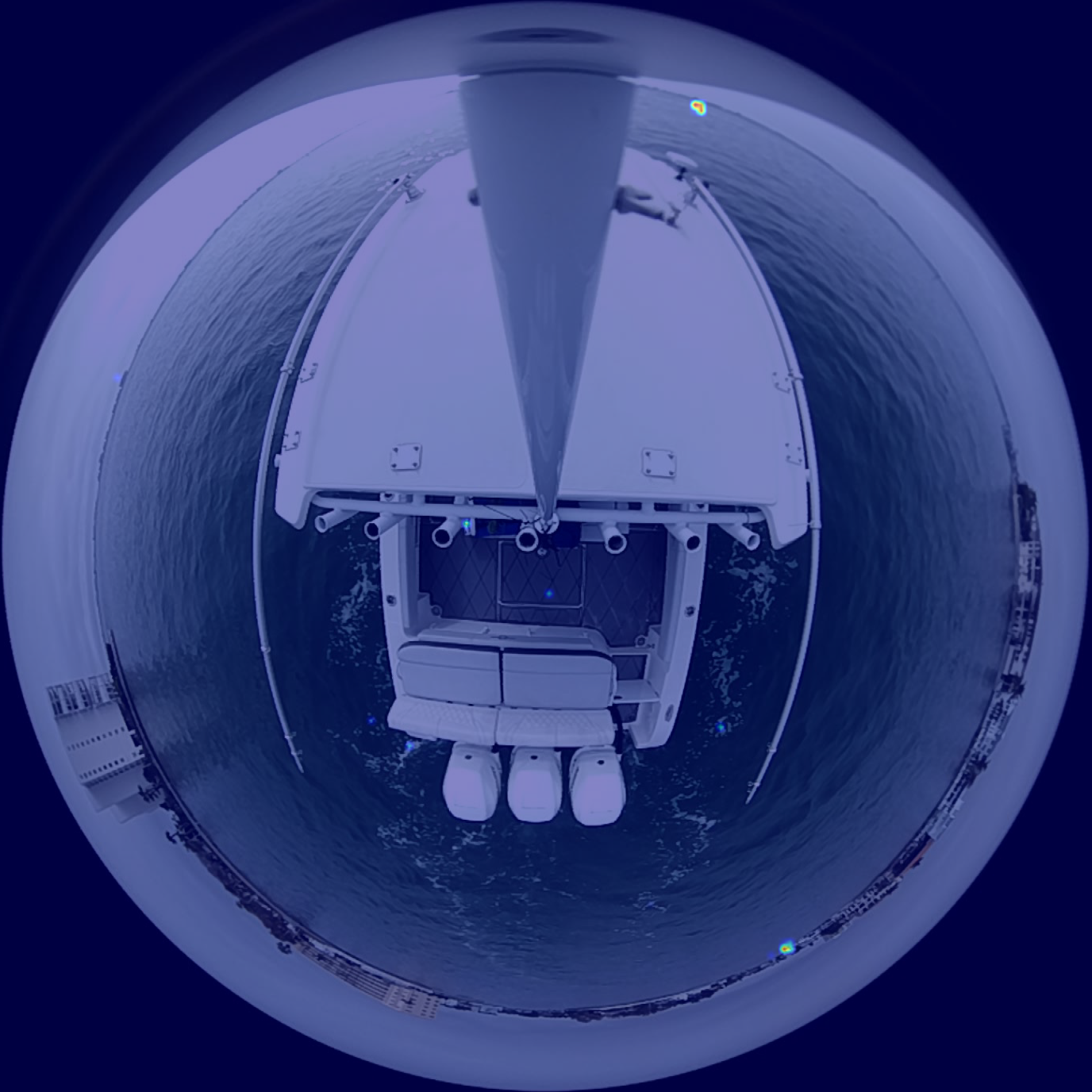} \\
        \small (a) Leaving boundary &
        \small (b) Occlusion &
        \small (c) Motion blur &
        \small (d) Sunlight glare \\
    \end{tabular}
    \caption{
        Qualitative comparison across challenging video scenarios: (a) leaving the image boundary, (b) occlusion, (c) motion blur, and (d) sunlight glare. 
        \textbf{Top row}: detection results comparing the single-frame baseline (green boxes) with our multi-frame YOLOv7 model (blue boxes). 
        \textbf{Middle row}: Grad-CAM++ heatmaps from the single-frame baseline. 
        \textbf{Bottom row}: heatmaps from our multi-frame model, showing improved focus and spatial understanding contributed by temporal context.
    }
    \label{fig:qualitative}
\end{figure*}
 
\subsubsection{Early Fusion vs. Grouped Convolution}

We investigated a variant where the first convolution layer applies grouped convolution across frames, separating early feature extraction. The intuition was to restrict low-level feature
extraction to each frame independently, only mixing temporal
information in later layers. 
 However, results in Table \ref{tab:grouped_conv} show consistent performance drops compared to standard early fusion.

This suggests that fusing spatial-temporal information early at the pixel-level allows the network to learn richer joint features, whereas restricting low-level channels by frame (grouping) hinders effective feature integration.

\begin{table}[ht]
\centering
\caption{Impact of grouped convolution at first layer on detection performance on MOT20Det.}
\label{tab:grouped_conv}
\begin{tabular}{lcccc}
\toprule
\textbf{YOLOv7-tiny}& \textbf{P} & \textbf{R} & \textbf{mAP@0.5} & \textbf{mAP@.5:.95} \\
\midrule
 3 frames adj. & \textbf{0.943} & 0.731 & \textbf{0.841} & \textbf{0.462} \\
3 frames adj., group & 0.920 & \textbf{0.746} & 0.829 & 0.444 \\
\midrule
 7 frames adj. & \textbf{0.946} & \textbf{0.755} & \textbf{0.855} &\textbf{0.478} \\
7 frames adj., group & 0.898 & 0.711 & 0.804 & 0.406 \\
\midrule
 3 frames, step 3 & \textbf{0.926} & \textbf{0.802 }&\textbf{ 0.873} &\textbf{0.490 }\\
3 frames, step 3, group & 0.925 & 0.736 & 0.829 & 0.448 \\
\bottomrule
\end{tabular}
\end{table}

\subsubsection{Scaling to Larger Models}

We evaluate the applicability of our approach on the larger YOLOv7 model, which offers  higher baseline accuracy compared to its tiny variant. As summarized in Table~\ref{tab:adjacent_frames_v7}, adding temporal context yields only marginal improvements over the already strong single-frame baseline. We hypothesize that the high-capacity YOLOv7 model already achieves strong single-frame performance, leaving less opportunity for additional temporal information to further improve detection. 

Notably, the 3-frame adjacent setup provides a slight boost in mAP@0.5:0.95 (+0.9), and sparse configurations such as 3-frame step-3 with grouped convolutions yield competitive results with minimal complexity increase.



Importantly, the multi-frame version of YOLOv7-tiny closes much of the performance gap to YOLOv7, highlighting the value of temporal modeling especially for compact models intended for deployment on resource-constrained devices.

\begin{table}[h]
\centering
\renewcommand{\arraystretch}{1.2}
\caption{Impact of different input settings on YOLOv7 performance on MOT20Det.}
\label{tab:adjacent_frames_v7}
\begin{tabular}{lcccc}
\toprule
\textbf{YOLOv7}& \textbf{P} & \textbf{R} & \textbf{mAP@0.5} & \textbf{mAP@.5:.95} \\
\midrule
\textit{1 frame (baseline)} & \textbf{0.964} & 0.779 & \textbf{0.882} & 0.511 \\
3 frames  adjacent & 0.954 & \textbf{0.796} & 0.879 & \textbf{0.520} \\
5 frames  adjacent & 0.953 & 0.747 & 0.852 & 0.473 \\
7 frames  adjacent & 0.955 & 0.774 & 0.867 & 0.497 \\
9 frames  adjacent & 0.850 & 0.697 & 0.778 & 0.411 \\
3 frames, step 3 & 0.953 & 0.757 & 0.855 & 0.493 \\
7 frames, step 3  & 0.957 & 0.729 & 0.849 & 0.487\\
3 frames, step 3, group  & 0.954 & \textbf{0.798} & 0.878 & \textbf{0.518} \\
7 frames, step 3, group  & 0.958 & 0.767 & 0.864 & 0.502\\
\bottomrule
\end{tabular}
\end{table}

\subsubsection{Computational Overhead}
An important consideration for deploying multi-frame detectors is the computational cost. 
Table~\ref{tab:efficiency_results} summarizes the number of parameters and GFLOPS on RTX3090 for various configurations. 

Notably, multi-frame models incur minimal increase in parameters: e.g., going from a 1-frame to 7-frame input on YOLOv7-tiny only grows the model from 6.006M to 6.011M parameters, by less than 0.1\%. 


Inference speed remains real-time, as reported in 
Table~\ref{tab:inference_speed}. Our YOLOv7-tiny multi-frame detector achieves 55 FPS on an NVIDIA Orin AGX, making it suitable for robotics deployments with strict latency constraints.

\begin{table}[ht]
\centering
\caption{Comparison of model size and computational cost for YOLOv7-tiny and YOLOv7 baselines with different multi-frame input configurations.}
\label{tab:efficiency_results}
\begin{tabular*}{\columnwidth}{@{\extracolsep{\fill}} l cc cc}
\toprule
\multirow{2}{*}{ } & \multicolumn{2}{c}{\textbf{YOLOv7-tiny}} & \multicolumn{2}{c}{\textbf{YOLOv7}} \\ 
\cmidrule(lr){2-3} \cmidrule(lr){4-5}
& Params & GFLOPS & Params & GFLOPS \\ 
\midrule
\textit{1 frame (baseline)} & 6,006 k& 13.0 & 36,479 k & 103.2\\
3 frames & 6,008 k & 13.4 & 36,481 k&104.6\\
5 frames & 6,010 k & 13.7 & 36,483 k&106.0\\
7 frames & 6,011 k & 14.1 & 36,485 k &107.4\\
9 frames & 6,013 k & 14.3 & 36,486 k&108.8\\
3 frames, group & 6,046 k & 15.3 & 36,520 k&112.1\\
7 frames, group & 6,049 k & 16.0 & 36,523 k&115.0\\
\bottomrule
\end{tabular*}
\end{table}


\begin{table}[t]
\centering
\caption{Inference speed (FPS) for 1024×1024 resolution images on NVIDIA Orin AGX.}
\label{tab:inference_speed}
\begin{tabular}{l cl}
\toprule
Model &   \textit{1 frame (baseline)}&Multi-frame\\
\midrule
YOLOv7-tiny& 55 &53-55\\
YOLOv7& 43 &41-43\\
\bottomrule
\end{tabular}
\end{table}

\subsubsection{Generalization to BOAT360}

We further evaluate on the BOAT360 dataset, characterized by moving cameras, fisheye distortion, and dynamic water scenes. 
As shown in Table~\ref{tab:adjacent_frames_BOAT360}, the best results are achieved with 3-frame input, achieving a notable improvement over the single-frame baseline for both YOLOv7-tiny (+12.8 mAP@0.5) and YOLOv7 (+8.1 mAP@0.5). 
However, with more than 5 frames the performance slightly degrades, likely due to rapid scene changes between distant frames. 

This indicates that our multi-frame framework generalizes well to new domains, and that moderate temporal context suffices when scene dynamics are fast.


\begin{table}[ht]
\centering
\caption{Impact of number of adjacent frames on detection performance on BOAT360.}
\label{tab:adjacent_frames_BOAT360}
\begin{tabular}{lcccc}
\toprule
\textbf{YOLOv7-tiny}& \textbf{P} & \textbf{R} & \textbf{mAP@0.5} & \textbf{mAP@.5:.95} \\
\midrule
\textit{1 frame (baseline)} & 0.653& 0.362& 0.411& 0.134\\
3 frames adjacent & 0.689&   \textbf{0.617}&   \textbf{0.539}& \textbf{0.215}\\
5 frames adjacent & 0.555& 0.532& 0.444& 0.146\\
7 frames adjacent & 0.616& \textbf{0.617}& 0.537& 0.186\\
9 frames adjacent &\textbf{0.674}&\textbf{0.617}& 0.526& 0.184\\
\midrule
\textbf{YOLOv7}& \textbf{ } & \textbf{ } & \textbf{ } & \textbf{ } \\
\midrule
\textit{1 frame (baseline)} & 0.729 & 0.574 & 0.583 & 0.234 \\
3 frames adjacent & 0.750          &   \textbf{0.702}       &   \textbf{0.664}& \textbf{0.277}\\
5 frames adjacent & 0.674& 0.66& 0.642& 0.25\\
7 frames adjacent & 0.577& 0.638& 0.523& 0.208\\
9 frames adjacent & \textbf{0.794}& 0.574& 0.606& 0.201\\
\bottomrule
\end{tabular}
\end{table}

Finally, to qualitatively validate our findings, we present visual comparisons in Fig.~\ref{fig:qualitative}. 
The examples showcase challenging scenarios where the multi-frame detector demonstrates clear advantages over the single-frame baseline, including partial object truncation at the image boundary, motion blur, occlusion, and sunlight glare. 
In these cases, the multi-frame model (blue boxes) exhibits superior robustness, detecting more objects correctly compared to the single-frame model (green boxes), highlighting its effectiveness in handling difficult real-world conditions. 
In addition to detection outputs, we visualize gradient-based class activation maps with Grad-CAM++ \cite{chattopadhay2018grad} for both the single-frame baseline and our multi-frame model. The activation maps are derived from the penultimate layer of the networks, as the final layer is specialized for producing detection heads and lacks meaningful spatial features. These maps reveal where the models focus their attention during inference. The baseline often produces diffuse activations in the presence of visual challenges. In contrast, the multi-frame model exhibits more localized attention aligned with object regions, suggesting stronger temporal awareness. These qualitative insights further highlight the benefit of incorporating multiple frames during training and inference.

\section{Conclusion}

In this work, we presented a simple yet effective strategy to enhance object detection by leveraging multi-frame temporal context within standard single-frame detector architectures. By stacking consecutive frames at the input level and supervising only the detection output corresponding to the latest frame, our approach enriches the early feature extraction process without introducing substantial complexity or computational overhead.

Through comprehensive experiments on MOT20Det and BOAT360 datasets, we demonstrated that lightweight models such as YOLOv7-tiny benefit significantly from temporal information, achieving up to 8\% relative improvement in detection accuracy. Our ablation studies revealed that moderate temporal context (e.g., 3 to 7 frames) and temporal step sampling are crucial for maximizing the performance gains, while early fusion of frames outperforms grouped convolutions. Furthermore, our approach generalizes well to challenging dynamic environments, as evidenced by its strong performance on the BOAT360 dataset.

Importantly, the proposed method scales efficiently, maintaining real-time inference speeds and introducing minimal parameter growth, making it suitable for deployment in robotics and mobile applications. Additionally, we contribute the BOAT360 dataset, which will be released to the community upon paper acceptance, to foster further research into robust video-based object detection under dynamic and non-ideal conditions.

In future work, we aim to extend this early fusion paradigm to other detection backbones, explore adaptive frame selection strategies, and investigate integration with lightweight temporal attention mechanisms to further boost video object detection performance while maintaining real-time constraints.

\bibliographystyle{./IEEEtran} 
\bibliography{./IEEEabrv,./IEEEexample}

\end{document}